\documentclass{article}  
\usepackage{xspace}
\newcommand{\eg}{e.g.\@\xspace}

\usepackage[preprint]{neurips_2025}

\usepackage[utf8]{inputenc} 
\usepackage[T1]{fontenc}    
\usepackage{hyperref}       
\usepackage{url}            
\usepackage{booktabs}       
\usepackage{amsfonts}       
\usepackage{nicefrac}       
\usepackage{microtype}      
\usepackage{xcolor}         
\usepackage{graphicx}
\usepackage{booktabs}
\usepackage{mathtools}   
\usepackage{algorithm}
\usepackage{algpseudocode}

\usepackage{cleveref}

\title{HiAR: Efficient Autoregressive Long Video Generation via Hierarchical Denoising}

\author{%
  Kai Zou$^{1,5}$ \quad
  Dian Zheng$^{2}$ \quad
  Hongbo Liu$^{3}$ \quad
  Tiankai Hang$^{4}$ \quad
  Bin Liu$^{1,5}$\textsuperscript{*} \quad
  Nenghai Yu$^{1,5}$ \\[0.5em]
  {\small $^1$University of Science and Technology of China \enspace
  $^2$The Chinese University of Hong Kong \enspace
  $^3$Tongji University}\\[0.2em]
  {\small $^4$Tencent Hunyuan \enspace
  $^5$Anhui Province Key Laboratory of Digital Security, USTC}\\[0.2em]
  {\small \texttt{kzou@mail.ustc.edu.cn}}, {\small \textsuperscript{*}Corresponding author.}\\[0.1em]
  \href{https://jacky-hate.github.io/HiAR/}{https://jacky-hate.github.io/HiAR/}
}

\begin{document}

\maketitle

\begin{abstract}
Autoregressive (AR) diffusion offers a promising framework for generating videos of theoretically infinite length. However, a major challenge is maintaining temporal continuity while preventing the progressive quality degradation caused by error accumulation. To ensure continuity, existing methods typically condition on highly denoised contexts; yet, this practice propagates prediction errors with high certainty, thereby exacerbating degradation. In this paper, we argue that a highly clean context is unnecessary. Drawing inspiration from bidirectional diffusion models, which denoise frames at a shared noise level while maintaining coherence, we propose that conditioning on context at the same noise level as the current block provides sufficient signal for temporal consistency while effectively mitigating error propagation.
Building on this insight, we propose \textbf{HiAR}, a hierarchical denoising framework that reverses the conventional generation order: instead of completing each block sequentially, it performs causal generation across all blocks at every denoising step, so that each block is always conditioned on context at the same noise level. This hierarchy naturally admits pipelined parallel inference, yielding a $\sim$1.8$\times$ wall-clock speedup in our 4-step setting. We further observe that self-rollout distillation under this paradigm amplifies a low-motion shortcut inherent to the mode-seeking reverse-KL objective. To counteract this, we introduce a forward-KL regulariser in bidirectional-attention mode, which preserves motion diversity for causal inference without interfering with the distillation loss. On VBench (20\,s generation), HiAR achieves the best overall score and the lowest temporal drift among all compared methods.
\end{abstract}

\section{Introduction}

\begin{figure}[tbh]
    \centering
    \includegraphics[width=1.0\linewidth]{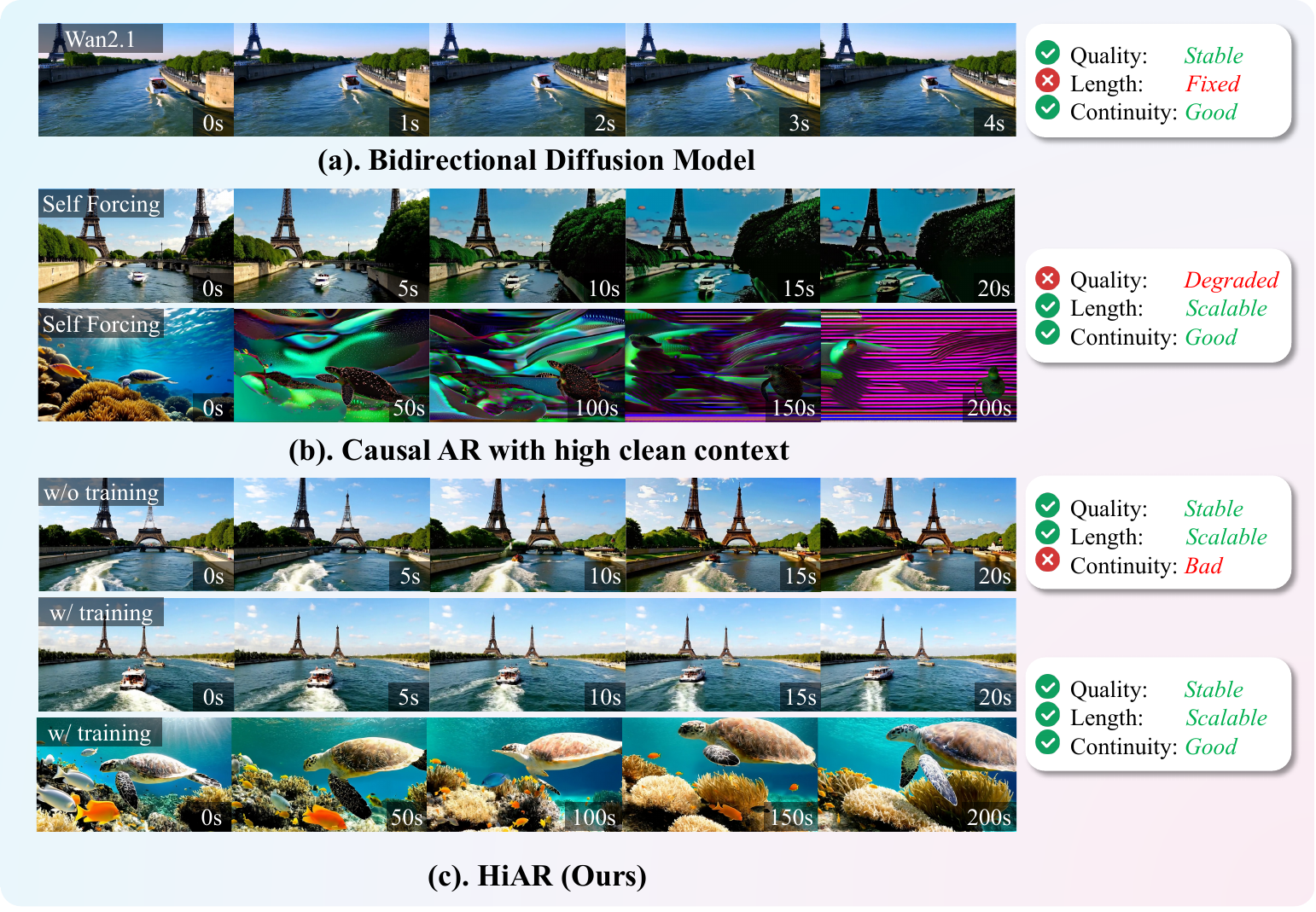}
    \vspace{-4mm}
    \caption{\textbf{Motivation.} (a)~Bidirectional diffusion (Wan2.1) proves that a shared noise level provides sufficient context for temporal coherence, though limited to a fixed horizon. (b)~Standard AR (Self-Forcing) scales length but suffers quality drift, as conditioning on fully clean context amplifies error propagation. (c)~Applying our hierarchical denoising (matched-noise context) only at inference (\emph{w/o training}) mitigates drift but breaks continuity due to train--test mismatch; \textbf{HiAR} retrains under the hierarchical pipeline (\emph{w/ training}), achieving scalable long-video generation with stable quality and seamless continuity.}
    \label{fig:teaser}
    \vspace{-3mm}
\end{figure}

\label{sec:intro}

Recent years have witnessed rapid progress in video generation, with Diffusion Transformer (DiT)~\cite{peebles2023scalable} backbones powering strong foundation models~\cite{ho2022video,blattmann2023stable,yang2024cogvideox,polyak2024moviegen,opensora2024,wan2025wan,sora2024} and conditional paradigms---including image-to-video and video-to-video---further broadening controllable generation.
A remaining frontier is long-horizon, and ultimately open-ended, video generation, central to interactive agents and world models~\cite{he2025matrixgame20opensourcerealtime,ye2025yanfoundationalinteractivevideo,mao2025yume15textcontrolledinteractiveworld,sun2025worldplaylongtermgeometricconsistency,hong2025relicinteractivevideoworld,tang2026hunyuangamecraft2instructionfollowinginteractivegame}.
To scale video duration, causal autoregressive (AR) generation~\cite{wu2025packforcememorylongform,pyramidflow2024,magi2025,skyreels2025} is increasingly attractive: it supports streaming output, indefinite extension, and real-time interaction.

However, a critical challenge in this pipeline is maintaining strict temporal continuity between consecutive video blocks while simultaneously preventing distribution drift (\eg, oversaturation, over-sharpening, motion repetition, and semantic drift) caused by error accumulation. To ensure temporal coherence, existing methods mainly denoise the previous frames into a \emph{highly clean} context before generating the next. Consequently, every denoising step of the current block is conditioned on a context with noise level $t_c=0$ (maximal SNR). While this highly clean context anchors the temporal consistency, it inadvertently causes the model to propagate accumulated prediction errors forward with high confidence, thereby exacerbating the degradation, as illustrated in \cref{fig:teaser}(b).

In this work, we recognize that \textbf{a highly clean context is not a prerequisite}. Drawing inspiration from bidirectional diffusion models, which denoise all frames concurrently from a shared noise level, yet still yield temporally coherent videos, as shown in \cref{fig:teaser}(a), demonstrating that noisy context already provides sufficient signal for continuity while reducing error propagation. 
Based on this principle, we introduce \textbf{HiAR}, a Hierarchical Denoising paradigm that swaps the denoising order: instead of fully denoising previous blocks first, we perform causal generation across all blocks within each denoising step, then move to the next step.
This simple yet fundamental change substantially reduces inter-block error transmission and improves long-horizon stability as shown in \cref{fig:teaser}(c, \emph{w/o training}).
Moreover, the hierarchical structure enables pipelined parallelism across denoising steps at inference time, improving wall-clock efficiency ($\times$1.8).

To maintain train--test consistency, we retrain under the hierarchical denoising pipeline.
However, we find that self-rollout distillation~\cite{selfforcing2025,yin2024improving} exhibits a low-motion shortcut that worsens over training—consistent with the mode-seeking tendency of DMD-style reverse-KL objectives~\cite{lu2025adversarial}: the model gradually collapses into near-static outputs that minimise distillation loss but sacrifice dynamics.
Hierarchical denoising amplifies this effect, as the increased learning difficulty of conditioning on multi-level noisy contexts requires more training steps.
Empirically, we find that motion diversity under bidirectional-attention denoising is strongly correlated with that under causal AR inference.
Motivated by this observation, we introduce a distillation-based forward-KL regulariser computed in bidirectional-attention denoising mode, effectively preventing dynamics collapse for the \emph{causal} inference path and enabling stable long-step training.

We conduct extensive evaluation on VBench~\cite{huang2024vbench,zheng2025vbench20advancingvideogeneration} and a dedicated drift metric tailored to long-horizon rollouts, together with thorough ablations, demonstrating that HiAR yields more stable long video generation and validating the contribution of each component. The visual result is shown in \cref{fig:teaser}(c, \emph{w/ training}).

We highlight the main contributions of this paper below:
\begin{itemize}
    \item We propose \textbf{HiAR}, a hierarchical denoising pipeline that performs causal generation across blocks within each denoising step, substantially reducing inter-block error transmission and enabling pipelined inference across hierarchy levels for $\sim$1.8$\times$ wall-clock speedup in our implementation.
    \item We introduce a simple forward-KL regulariser via bidirectional-attention distillation to prevent low-motion shortcuts in self-rollout training, enabling stable scaling to long training schedules while preserving dynamics.
    \item Extensive experiments on VBench and a dedicated drift metric, together with thorough ablations, demonstrate the long-horizon stability and the effectiveness of each component.
\end{itemize}

\section{Background}
\label{sec:background}

\subsection{Diffusion Models and Flow Matching}
\label{sec:bg_diffusion}

Diffusion-based generative models~\cite{ho2020denoising,song2021score} learn to reverse a forward noising process that gradually corrupts data into Gaussian noise.
In this work we adopt the flow matching  formulation~\cite{lipman2023flow,liu2023flow,albergo2023building}. 
Let $x_0\sim p_{\text{data}}$ denote a clean data sample and $\epsilon\sim\mathcal{N}(0,I)$ standard Gaussian noise.
The forward interpolation (corruption) at continuous time $t\in[0,1]$ is defined as
\begin{equation}\label{eq:xt}
  x_t \;=\; (1-\sigma_t)\,x_0 \;+\; \sigma_t\,\epsilon,
  \qquad
  \sigma_t \;=\; \frac{s\cdot t}{1+(s-1)\cdot t},
\end{equation}
where $s>0$ is a shift parameter that controls the noise schedule curvature.
At $t=0$ we recover $x_0$; at $t=1$ we obtain (approximately) pure noise.
A neural network $v_\theta(x_t,t)$ is trained to predict the velocity field
\begin{equation}\label{eq:velocity}
  v^{*}(x_t,t) \;=\; \epsilon - x_0,
\end{equation}
so that clean data can be recovered by integrating the probability-flow ODE backward from $t=1$ to $t=0$.
In practice, one discretises the trajectory into $S$ steps $1=t_1>t_2>\cdots>t_S>0$ and applies the Euler update
\begin{equation}\label{eq:euler}
  x_{t_{j+1}} \;=\; x_{t_j} \;+\; v_\theta(x_{t_j},t_j)\,\bigl(\sigma_{t_{j+1}}-\sigma_{t_j}\bigr).
\end{equation}

\subsection{Autoregressive Video Diffusion}
\label{sec:bg_ar}

Bidirectional-attention diffusion models~\cite{sora,wan2025,kling2.6,veo3,gen4} operate on a fixed temporal window and cannot easily scale to arbitrary durations.
Causal AR generation~\cite{po2025bagger,liu2025rolling,lu2025reward,zhang2025test,yang2025longlive,lin2025autoregressive} overcomes this limitation by generating frames in a streaming manner: it naturally supports indefinite extension, allows real-time intervention, and provides a principled interface for interactive control---making it a key building block toward world models~\cite{he2025matrixgame20opensourcerealtime, ye2025yanfoundationalinteractivevideo,mao2025yume15textcontrolledinteractiveworld}. To generate videos beyond a fixed temporal window, recent work partitions the video latent sequence into $N$ successive blocks $\{B_1,\ldots,B_N\}$, each containing $k$ frames, and generates them autoregressively: for $n=2,\ldots,N$, block $B_n$ is denoised conditioned on the previously generated blocks $B_{<n}$.

Concretely, let $x_t^{(n)}$ denote the noisy latent of block $n$ at timestep $t$.
The denoiser is queried as
\begin{equation}\label{eq:ar_denoise}
  v_\theta\!\bigl(x_t^{(n)},\,t \;\big|\; c_{<n}\bigr),
\end{equation}
where $c_{<n}$ is the context representation of blocks $B_1,\ldots,B_{n-1}$ injected through causal attention: the query tokens come from $x_t^{(n)}$ while the key/value tokens include $c_{<n}$.

Under teacher forcing~\cite{williams1989learning,gao2024ca2,hu2024acdit,jin2024pyramidal,zhang2025test}, training conditions on ground-truth context ($c_{<n}=x_0^{(<n)}$), whereas at inference $c_{<n}$ consists of model predictions $\hat{x}_0^{(<n)}$; this train--test mismatch causes per-step errors to accumulate along the autoregressive chain---exposure bias~\cite{bengio2015scheduled}---manifesting as progressive over-saturation, motion repetition, and semantic drift, collectively termed distribution drift.
Diffusion Forcing~\cite{chen2024diffusion,yin2025slow,chen2025skyreels,gu2025long,teng2025magi,song2025history,po2025bagger} mitigates this by training with independent per-token noise levels, so the model learns to denoise under heterogeneous noise conditions and gains robustness to partially noisy contexts at inference.

Self-Forcing~\cite{selfforcing2025,yin2024improved,yin2024one,yi2025deep} further closes the train--test gap through self-rollout training: during each training iteration, a block is first rolled out with the student model $v_\theta$ to obtain $\hat{x}_0^{(n-1)}$, which is then used as context for the next block's denoising.
The training objective is an asymmetric Distribution Matching Distillation (DMD) loss~\cite{yin2024improving,yin2024onestepdiffusiondistributionmatching}, formulated as a reverse KL divergence between the student's one-step output distribution and the teacher's multi-step output distribution:
\begin{equation}\label{eq:dmd}
  \mathcal{L}_{\text{DMD}} \;=\; \mathbb{E}_{t,\,x_t}\!\Big[\,D_{\mathrm{KL}}\!\big(\,p_\theta(x_0\mid x_t)\;\|\;p_{\text{teacher}}(x_0\mid x_t)\,\big)\,\Big],
\end{equation}
where $p_\theta(x_0\mid x_t)$ denotes the distribution induced by the student's single Euler step from $x_t$, and $p_{\text{teacher}}$ is the distribution obtained by multi-step ODE integration with the teacher model.
This reverse KL encourages the student to mode-seek toward the teacher's high-density regions.
In practice the gradient is computed via a learned score difference between student and teacher distributions.
While Self-Forcing achieves notable improvements at moderate horizons, it still employs $t_c=0$ for context (i.e., the predicted clean frame), propagating errors with maximal confidence.

\section{Method}
\label{sec:method}

\begin{figure}[tb]
  \centering
  \includegraphics[width=1\linewidth]{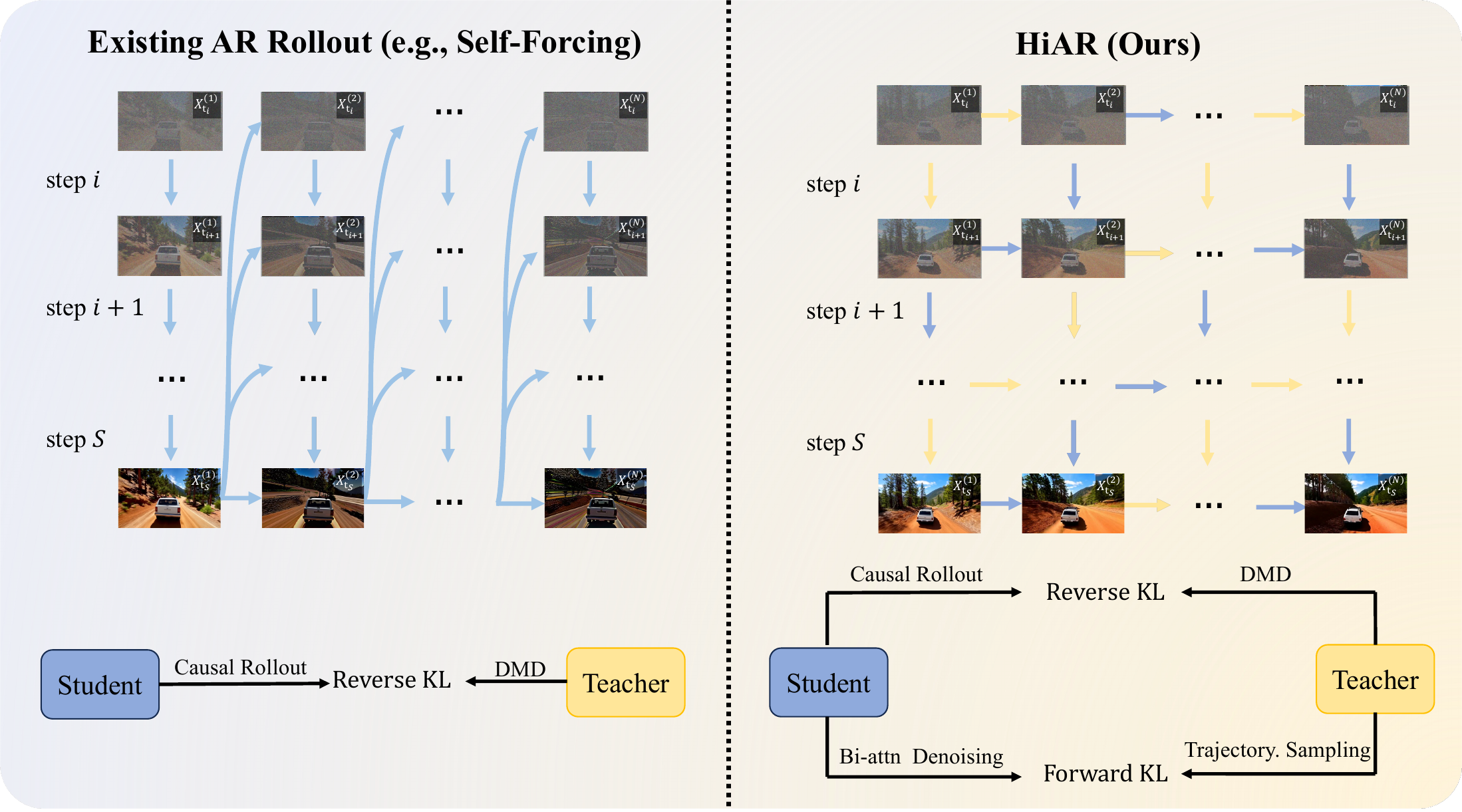}

\caption{\textbf{Overview of HiAR.} \textbf{Left:} Existing block-first AR (\eg, Self-Forcing) fully denoises each block before generating the next, conditioning every step on predicted clean context and thus amplifying inter-block error propagation. \textbf{Right:} Our hierarchical denoising performs causal generation across all blocks within each denoising step, conditioning on context at the matched noise level to suppress error accumulation. \textbf{Bottom:} Training combines causal self-rollout with a reverse-KL (DMD) loss for distillation, and a forward-KL regulariser computed in bidirectional-attention mode via teacher trajectory sampling to preserve motion diversity.}
  \label{fig:overview}
  \vspace{-3mm}
\end{figure}

We now formalise the intuition developed in Sec.~\ref{sec:intro}: the context noise level $t_c$ governs a bias--information trade-off, and the optimal choice is $t_c^*=t_{j+1}$---the output noise level of the current denoising step.
We first derive this result analytically and then build upon it to design Hierarchical Denoising.

\subsection{Context Noise Level and Error Propagation}
\label{sec:method_tc}

\noindent\textbf{Error decomposition.}
Consider block $B_n$ being denoised at step $j$ (from noise level $t_j$ to $t_{j+1}$).
Let $x_0^{(n-1)}$ denote the ground-truth clean latent of the preceding block and $\hat{x}_0^{(n-1)}=x_0^{(n-1)}+\delta^{(n-1)}$ the model's prediction, where $\delta^{(n-1)}$ is the accumulated prediction error.
In AR diffusion, the context for $B_n$ is derived from $\hat{x}_0^{(n-1)}$ and presented at some noise level $t_c\in[0,1]$:
\begin{equation}\label{eq:context}
  c_{n-1}^{(t_c)}
  \;=\; (1-\sigma_{t_c})\,\hat{x}_0^{(n-1)} + \sigma_{t_c}\,\eta,
  \quad \eta\sim\mathcal{N}(0,I).
\end{equation}
Expanding $\hat{x}_0^{(n-1)}$ decomposes the context into three terms:
\begin{equation}\label{eq:three_term}
  c_{n-1}^{(t_c)}
  \;=\;
  \underbrace{(1{-}\sigma_{t_c})\,x_0^{(n-1)}}_{\text{true signal}}
  \;+\;
  \underbrace{(1{-}\sigma_{t_c})\,\delta^{(n-1)}}_{\text{propagated bias}}
  \;+\;
  \underbrace{\sigma_{t_c}\,\eta}_{\text{stochastic perturbation}}.
\end{equation}
The true-signal and propagated-bias terms share the same coefficient $(1{-}\sigma_{t_c})$, while the stochastic term carries the complementary coefficient $\sigma_{t_c}$.
The noise level $t_c$ thus controls a \emph{bias--information trade-off}: raising $t_c$ attenuates the bias but simultaneously reduces the useful conditioning signal by the same factor.
In particular, prior AR methods~\cite{selfforcing2025} use $t_c=0$, which reduces Eq.~\ref{eq:three_term} to $c_{n-1}^{(0)}=x_0^{(n-1)}+\delta^{(n-1)}$ and propagates the full prediction error with no attenuation.

\noindent\textbf{Temporal causality.}
To produce temporally coherent continuations, the context must carry at least as much information as the current block possesses after step $j$.
Under Eq.~\ref{eq:xt}, the signal-to-noise ratio $\mathrm{SNR}(t)={(1{-}\sigma_t)^2}/{\sigma_t^2}$ increases monotonically as $t$ decreases, so after step $j$ the current block at $t_{j+1}$ contains strictly more information than at $t_j$.
Temporal causality therefore requires
\begin{equation}\label{eq:tc_lb}
  \mathrm{SNR}(t_c) \;\geq\; \mathrm{SNR}(t_{j+1})
  \quad\Longleftrightarrow\quad
  t_c \;\leq\; t_{j+1}.
\end{equation}
Any $t_c$ satisfying this bound provides sufficient information for step $j$.
Since the bias coefficient $(1{-}\sigma_{t_c})$ decreases monotonically in $t_c$, choosing $t_c < t_{j+1}$ only transmits more prediction error without additional benefit.
The optimum is therefore the boundary of the constraint:
\begin{equation}\label{eq:tc_opt}
  \boxed{t_c^* \;=\; t_{j+1},}
\end{equation}
the noisiest context level that still fulfills temporal causality--- attenuating inter-block bias while retaining all information the denoiser needs at step $j$.

\subsection{Hierarchical Denoising}
\label{sec:method_hier}

The analysis above motivates a simple but fundamental change to the autoregressive denoising pipeline:
instead of fully denoising each block before moving to the next, we perform causal generation across all blocks at each denoising step.
We call this \emph{Hierarchical Denoising} (Fig.~\ref{fig:overview}).

\noindent\textbf{Inference procedure.}
The complete procedure is summarised in Alg.~\ref{alg:hier}.
At each step $j$, block $B_n$ is denoised with blocks $B_{<n}$ at noise level $t_{j+1}$ as context---the noisiest level that still preserves temporal causality (Sec.~\ref{sec:method_tc}).

\begin{algorithm}[tb]
\caption{Hierarchical Denoising}\label{alg:hier}
\begin{algorithmic}[1]
\Require Schedule $t_1{>}t_2{>}\cdots{>}t_S{\approx}0$;\; initial noise $\{x_{t_1}^{(n)}\}_{n=1}^N$
\Ensure  Generated blocks $\{\hat{x}_0^{(n)}\}_{n=1}^N$
\For{$j=1,\ldots,S$} \Comment{denoising steps}
  \For{$n=1,\ldots,N$} \Comment{causal block sweep with KV cache}
    \State $x_{t_{j+1}}^{(n)} \gets x_{t_j}^{(n)} + v_\theta\!\bigl(x_{t_j}^{(n)},t_j \mid x_{t_{j+1}}^{(<n)}\bigr)\,(\sigma_{t_{j+1}}-\sigma_{t_j})$
  \EndFor
  \State Update KV cache with $\{x_{t_{j+1}}^{(n)}\}_{n=1}^N$
\EndFor
\State \Return $\hat{x}_0^{(n)} \gets x_{t_S}^{(n)}$ for $n=1,\ldots,N$
\end{algorithmic}
\end{algorithm}

\noindent\textbf{Pipelined parallelism.}
In Alg.~\ref{alg:hier}, block $B_n$ at step $j$ depends only on $B_{<n}$ at step $j$ and on $B_n$ at step $j{-}1$, so blocks at different $(n,j)$ positions that lie on the same anti-diagonal of the $N{\times}S$ grid are mutually independent.
We exploit this by assigning each denoising step to a dedicated process and traversing the grid along its $N{+}S{-}1$ anti-diagonals, with inter-stage latents exchanged via asynchronous point-to-point communication.
Within each stage, na\"ively updating the KV cache for block $B_n$ and denoising block $B_{n+1}$ are two separate forward passes, totalling $2N$ per stage.
We observe that under causal attention the two operations can be fused into one forward call by concatenating $[c^{(n)},\, x_{t_j}^{(n+1)}]$ along the frame dimension with per-frame timesteps $[t_{j+1},\ldots,t_{j+1},\, t_j,\ldots,t_j]$: the first segment writes $B_n$'s context into the KV cache while the second segment denoises $B_{n+1}$ attending to the freshly written keys and values.
This fusion reduces the cost to $N{+}2$ passes per stage (one standalone denoise for the first block, $N{-}1$ fused passes, and one trailing cache write), yielding an overall ${\sim}1.8{\times}$ wall-clock speedup in our 4-step setting.

\subsection{Training with Forward-KL Regulation}
\label{sec:method_fkl}

Although hierarchical denoising already mitigates degradation at test time, a train--test gap remains when the model has been trained under the conventional block-first rollout.
We therefore retrain with self-rollout under the hierarchical schedule, optimising the DMD reverse-KL objective (Eq.~\ref{eq:dmd}) following Self-Forcing~\cite{selfforcing2025}. The overall training pipeline is illustrated in \cref{fig:overview} (bottom).

\noindent\textbf{The low-motion shortcut.}
As training progresses, temporal coherence improves yet motion diversity collapses: the model increasingly produces near-static videos.
The root cause is the mode-seeking nature of the reverse-KL objective: $D_{\mathrm{KL}}(p_\theta\|p_{\text{teacher}})$ is minimised when the student concentrates its mass on a single high-density mode, so it can reduce loss by generating low-motion outputs that are inherently easier to denoise and less prone to rollout errors.
Hierarchical denoising amplifies this shortcut, because conditioning on contexts at varying noise levels---rather than only clean ones---increases learning difficulty and demands more training steps, giving the mode-seeking objective more iterations to collapse onto the low-motion mode.

\noindent\textbf{Forward-KL regularisation via distillation.}
To counteract this shortcut, we introduce a complementary loss that penalises mode dropping.
We first run the teacher for a large number of ODE steps to obtain a dense denoising trajectory, from which we extract checkpoints $\{x_{t_1}^{\text{ref}},\ldots,x_{t_S}^{\text{ref}}\}$ aligned with the student's $S$-step schedule.
The student is then supervised to match each consecutive pair via a single Euler step:
\begin{equation}\label{eq:fkl}
  \mathcal{L}_{\text{FKL}}
  \;=\;
  \mathbb{E}_{i}\!\Big[\,
    \big\|\,
      v_\theta(x_{t_i}^{\text{ref}},\, t_i)
      \,-\,
      \tfrac{x_{t_{i+1}}^{\text{ref}} - x_{t_i}^{\text{ref}}}{\sigma_{t_{i+1}} - \sigma_{t_i}}
    \big\|^2
  \,\Big].
\end{equation}
Because the targets $x_t^{\text{ref}}$ are drawn from the teacher's distribution, optimising Eq.~\ref{eq:fkl} amounts to minimising a forward-KL-direction objective that encourages the student to cover the teacher's output modes rather than mode-seek, thereby preserving motion diversity.

\noindent\textbf{Decoupling from DMD.}
To prevent interference between $\mathcal{L}_{\text{FKL}}$ and the DMD objective, we adopt two design choices:
\begin{enumerate}
  \item \textbf{Bidirectional-attention mode only.}
  Motion dynamics under bidirectional and causal attention are strongly positively correlated (Sec.~\ref{sec:exp_ablation}), so regularising the former effectively constrains the latter.
  We therefore compute $\mathcal{L}_{\text{FKL}}$ exclusively in bidirectional-attention mode, leaving the causal self-rollout DMD loss unmodified and minimising gradient interference.

  \item \textbf{Early-step restriction.}
  Motion dynamics are governed by low-frequency structures established during the earliest denoising steps.
  We thus apply $\mathcal{L}_{\text{FKL}}$ only to the first $K$ of $S$ steps, leaving subsequent high-frequency refinement steps unconstrained.
\end{enumerate}
The overall training objective is
\begin{equation}\label{eq:total_loss}
  \mathcal{L}
  \;=\;
  \mathcal{L}_{\text{DMD}}
  \;+\;
  \lambda\,\mathcal{L}_{\text{FKL}},
\end{equation}
where $\lambda>0$ balances the two terms.
We ablate the choice of $K$ and the attention-mode decoupling strategy in Sec.~\ref{sec:exp_ablation}.

\section{Experiments}
\label{sec:exp}

\subsection{Setups}
\label{sec:exp_setup}

\begin{table}[thb]
  \centering
  \caption{\textbf{Quantitative comparison on 20\,s generation.} Throughput is in frames/s; Latency is in seconds; VBench scores (Total/Quality/Semantic/Dynamic) are on a 0--1 scale; Drift is our proposed drift metric. ``--'' indicates the model is non-autoregressive and drift is not applicable. Best distilled AR results are \textbf{bolded}.}
  \vspace{-2mm}
  \label{tab:main}
  \resizebox{0.95\linewidth}{!}{%
  \begin{tabular}{@{}l cc cccc c@{}}
    \toprule[0.12em]
    Model & Thru. & Lat. & Total$\uparrow$ & Quality$\uparrow$ & Semantic$\uparrow$ & Dynamic$\uparrow$ & Drift$\downarrow$ \\
    \midrule[0.1em]
    \multicolumn{8}{l}{\emph{Bidirectional video diffusion models}} \\
    LTX-Video~\cite{ltxvideo2025} & 8.98 & 13.5 & 0.766 & 0.789 & 0.685 & 0.458 & -- \\
    Wan2.1-1.3B~\cite{wan2025wan} & 0.78 & 103 & 0.802 & 0.813 & 0.766 & \textbf{0.690} & -- \\
    \midrule[0.1em]
    \multicolumn{8}{l}{\emph{Autoregressive video diffusion models}} \\
    NOVA~\cite{nova2025} & 0.88 & 4.1 & 0.773 & 0.777 & 0.757 & 0.444 & -- \\
    Pyramid Flow~\cite{pyramidflow2024} & 6.70 & 2.5 & 0.775 & 0.804 & 0.670 & 0.161 & -- \\
    SkyReels-V2-1.3B~\cite{skyreels2025} & 0.49 & 112 & 0.788 & 0.808 & 0.707 & 0.333 & -- \\
    MAGI-1-4.5B~\cite{magi2025} & 0.19 & 282 & 0.757 & 0.785 & 0.647 & 0.486 & -- \\
    \midrule[0.1em]
    \multicolumn{8}{l}{\emph{Distilled autoregressive video models}} \\
    CausVid~\cite{yin2025causvid} & 17 & 0.69 & 0.764 & 0.771 & 0.740 & 0.621 & 0.842 \\
    Self-Forcing~\cite{selfforcing2025} & 17 & 0.69 & 0.805 & 0.829 & 0.708 & 0.542 & 0.355 \\
    Causal Forcing~\cite{causalforcing2025} & 17 & 0.69 & 0.810 & 0.837 & 0.701 & 0.672 & 0.615 \\
    \textbf{HiAR (Ours)} & \textbf{30} & \textbf{0.30} & \textbf{0.821} & \textbf{0.846} & \textbf{0.723} & 0.686 & \textbf{0.257} \\
    \bottomrule[0.1em]
  \end{tabular}%
  }
\end{table}

\noindent\textbf{Implementation details.}
We use the Wan2.1-1.3B backbone~\cite{wan2025wan} as our base model. Following Self-Forcing~\cite{selfforcing2025}, we fine-tune the model with causal attention masking on 16k ODE solution pairs sampled from the base model.
We adopt a 4-step denoising schedule ($S=4$) and use Wan2.1-14B as the teacher model for the DMD critic.
All methods are implemented in a \emph{chunk-wise} manner, where each chunk contains 3 latent frames.
For the forward-KL regulariser, we sample 20\,k denoising trajectories (50 ODE steps each) from the Wan2.1-1.3B base model, and restrict $\mathcal{L}_{\text{FKL}}$ to the first denoising step only ($K=1$), with a balancing weight $\lambda=0.1$.
The critic model and generator are updated at a 5:1 ratio.
We train with a learning rate of $2{\times}10^{-6}$ and a total batch size of 64 for 20\,k steps on 5-second clips.
At inference time, we employ a sliding-window KV cache with a constant attention window of 5\,s.

\noindent\textbf{Evaluation metrics.}
We adopt the VBench~\cite{huang2024vbench} protocol, which measures 16 dimensions grouped into a Quality score and a Semantic score, providing a comprehensive assessment of average generation quality.
All models are sampled to 20\,s to evaluate long-video capability.
To quantify temporal degradation beyond aggregate scores, we introduce a drift metric suite specifically designed for long-horizon evaluation.
Each 20-second video is evenly divided into five temporal segments, and the following per-segment statistics are computed:
perceptual quality via MUSIQ~\cite{ke2021musiq} and CLIP-IQA~\cite{wang2023clipiqa};
temporal coherence via DINOv2~\cite{oquab2024dinov2} consecutive-frame cosine similarity and LPIPS~\cite{zhang2018lpips} consecutive-frame distance;
and low-level statistics including HSV saturation mean and Laplacian variance (sharpness).
For each metric, we report the slope of a linear fit over the five segments as a measure of drift rate.
All per-metric slopes are then normalised and aggregated via a weighted sum into a single \emph{Drift Score} (lower is better) that summarises overall temporal stability.

\noindent\textbf{Baselines.}
We compare against recent open-source video generation methods spanning three categories:
(i)~bidirectional diffusion models---\textbf{LTX-Video}~\cite{ltxvideo2025} (real-time Video-VAE + spatiotemporal transformer) and \textbf{Wan2.1-1.3B}~\cite{wan2025wan} (the foundation model shared by all distilled methods below);
(ii)~autoregressive diffusion models---\textbf{NOVA}~\cite{nova2025} (non-quantised temporal AR with spatial diffusion), \textbf{Pyramid Flow}~\cite{pyramidflow2024} (pyramidal flow matching with temporal pyramid), \textbf{SkyReels-V2}~\cite{skyreels2025} (1.3B; diffusion forcing with non-decreasing noise schedules), and \textbf{MAGI-1}~\cite{magi2025} (4.5B; block-causal attention);
(iii)~distilled AR models, all distilling Wan2.1 into a 4-step causal generator---\textbf{CausVid}~\cite{yin2025causvid} (bidirectional-to-AR DMD distillation), \textbf{Self-Forcing}~\cite{selfforcing2025} (self-rollout DMD training), and \textbf{Causal Forcing}~\cite{causalforcing2025} (use diffusion forcing as mid-training before self-rollout distillation).
All baselines use official checkpoints and are evaluated under identical prompts and generation lengths (5s for bidirectional models).

\subsection{Quantitative Results}

Table~\ref{tab:main} reports VBench scores, drift, and inference efficiency for all methods.

\noindent\textbf{VBench results.}
HiAR achieves the highest Total score (0.821) among all methods, surpassing both bidirectional and autoregressive baselines.
Notably, it attains the best Quality score (0.846) while maintaining a strong Semantic score (0.723), indicating that hierarchical denoising does not sacrifice semantic fidelity for visual quality.
On the Dynamic dimension, HiAR scores 0.686, closely preserving the motion diversity of the bidirectional Wan2.1-1.3B teacher (0.690) and substantially outperforming all other AR methods---including Causal Forcing (0.672) and Self-Forcing (0.542)---demonstrating the effectiveness of our forward-KL regulariser in preventing motion collapse.

\noindent\textbf{Temporal stability.}
On our proposed Drift metric, HiAR achieves 0.257, the lowest among all distilled AR models, indicating minimal quality degradation over the 20\,s horizon.
By contrast, CausVid exhibits the highest drift (0.842), consistent with its visible colour oversaturation at later segments; Self-Forcing (0.355) and Causal Forcing (0.615) show intermediate degradation.
HiAR reduces drift by 27.6\% relative to Self-Forcing (0.257 vs.\ 0.355), confirming that hierarchical denoising with matched context noise levels substantially mitigates the compounding inter-block error that drives long-horizon degradation.

\noindent\textbf{Inference efficiency.}
Owing to pipelined parallelism across hierarchy levels, HiAR achieves 30\,fps throughput and 0.30\,s per-chunk latency---a $\sim$1.8$\times$ wall-clock speedup over other distilled AR models (17\,fps, 0.69\,s) that share the same Wan2.1-1.3B backbone and 4-step denoising schedule.
This speedup comes at no cost to generation quality; in fact, HiAR simultaneously achieves the best VBench scores and the lowest drift.

\begin{figure}[th!]
    \centering
    \includegraphics[width=1.0\linewidth]{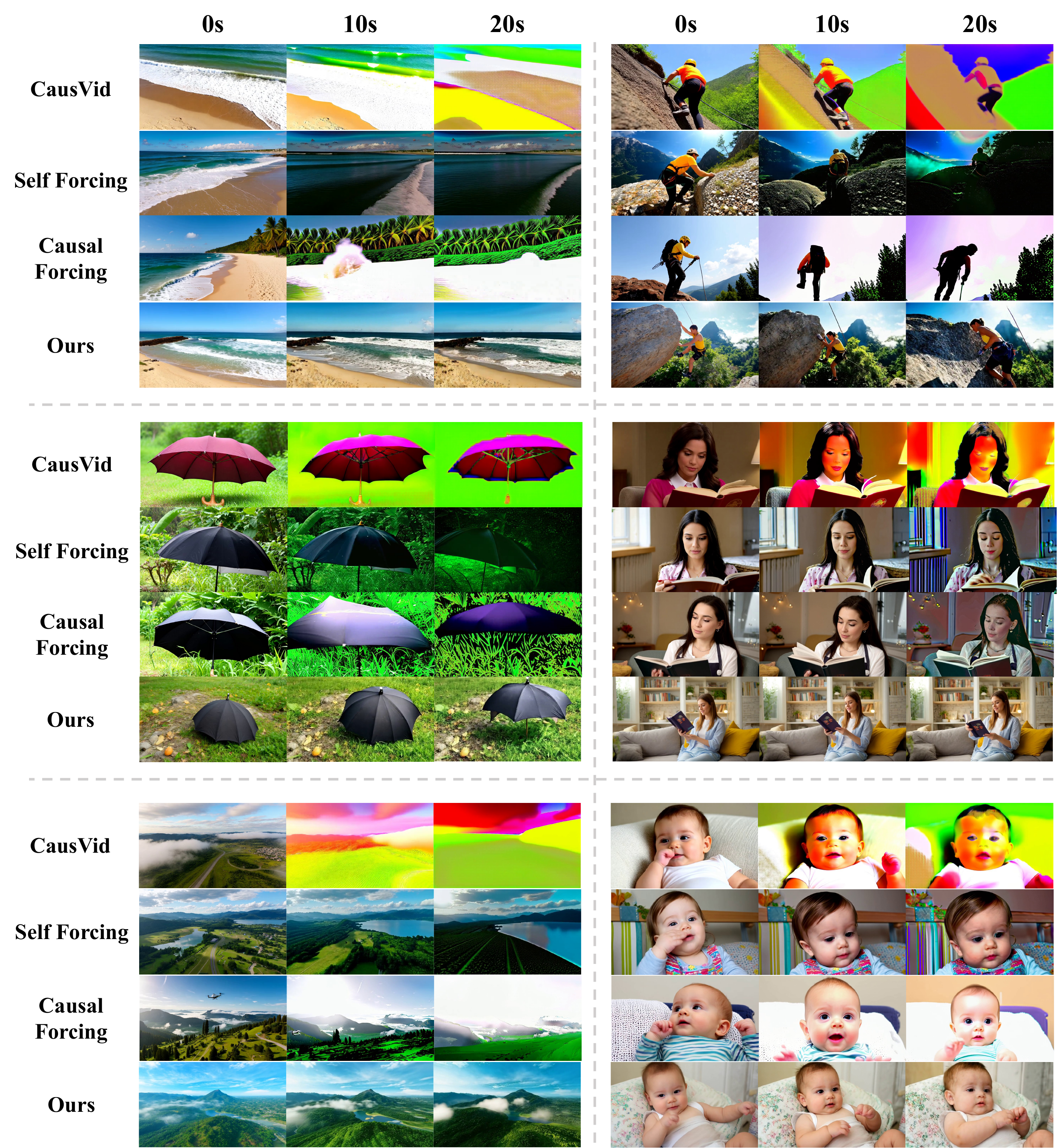}
    \vspace{-5mm}
    \caption{\textbf{Qualitative comparison of distilled AR models at 20\,s.} We show temporally sampled frames from six diverse prompts covering natural scenery, objects, and human subjects. HiAR maintains consistent colour and detail throughout, while baselines exhibit progressive degradation.
    }
    \vspace{-1mm}
    \label{fig:quality}
\end{figure}

\subsection{Qualitative Results}

Fig.~\ref{fig:quality} presents visual comparisons among all distilled autoregressive models on 20\,s generation across six diverse prompts spanning natural scenery (beach, mountain landscape), objects (umbrellas), and human subjects (rock climbing, woman reading, baby portrait).

CausVid exhibits the most severe degradation: frames progressively shift toward neon green and yellow tints, with scene content largely unrecognisable by 20\,s.
Self-Forcing and Causal Forcing alleviate this to some extent, yet still develop visible colour oversaturation and hue drift over time.
The degradation is particularly pronounced on human-centric content---facial regions suffer from unnatural colour casts and loss of fine detail (e.g., skin texture, facial features), which are perceptually salient and difficult to mask.
By contrast, HiAR maintains stable colour fidelity, sharpness, and structural coherence from the first frame to the last across all content types, with no perceptible drift in either scenery or portrait prompts.

\subsection{Ablation Studies}
\label{sec:exp_ablation}

We conduct ablations along two axes: the context noise level $t_c$ (Table~\ref{tab:abl_tc}) and the design choices of the forward-KL regulariser (Table~\ref{tab:abl_fkl}).
All variants are retrained under the same rollout mode used at inference to ensure train--test consistency, unless stated otherwise.

\noindent\textbf{Context noise level.}
Table~\ref{tab:abl_tc} compares three context noise configurations.
We evaluate overall video quality (Quality, Semantic), temporal smoothness approximated by the VBench motion smoothness score, and long-horizon stability (Drift).

\begin{table}[tb]
  \centering
  \caption{\textbf{Ablation study on context noise level $t_c$.} Quality, Semantic, and Smooth are VBench sub-scores; Drift is our proposed drift metric.}
  \label{tab:abl_tc}
  \begin{tabular}{@{}l cccc@{}}
    \toprule
    Context noise & Quality$\uparrow$ & Semantic$\uparrow$ & Smooth.$\uparrow$ & Drift$\downarrow$ \\
    \midrule
    $t_c=t_j$ (input level)           & 0.799 & 0.692 & 0.978 & \textbf{0.184} \\
    $t_c=t_{j+1}$ (output level; default) & \textbf{0.846} & \textbf{0.723} & 0.988 & 0.257 \\
    $t_c=0$ (Self-Forcing) & 0.829 & 0.708 & \textbf{0.991} & 0.355 \\
    \bottomrule
  \end{tabular}
\end{table}

When $t_c=t_j$ (the input noise level of the current step), the context carries the same noise level as the current block's input, meaning that block $B_n$ cannot observe the result of denoising step $j$ on block $B_{n-1}$---effectively removing intra-step causality.
While this yields the lowest drift (0.184), the lack of any one-step-ahead information substantially degrades generation quality (Quality 0.799 vs.\ 0.846) and produces noticeably unsmooth motion (Smooth.\ 0.978).
At the other extreme, $t_c=0$ (the standard Self-Forcing setting) fully denoises the context, exposing the model to maximum error propagation and the highest drift (0.355).
Our default $t_c=t_{j+1}$ (the output noise level)---where each block conditions on the context that has been denoised through the current step---strikes the optimal balance: it preserves nearly the same temporal smoothness as Self-Forcing (0.988 vs.\ 0.991) while substantially reducing drift and improving overall quality.

\noindent\textbf{Forward-KL regulation design.}
Table~\ref{tab:abl_fkl} ablates the attention mode, number of constrained denoising steps $K$, and the necessity of each component.
We focus on motion dynamics (Dynamic), overall quality (Quality, Semantic), and drift.

\begin{table}[tb]
  \centering
  \caption{\textbf{Ablation on forward-KL regulariser design.} ``bi-attn'', ``causal'' denotes the attention mode used for $\mathcal{L}_{\text{FKL}}$; ``$K$step'' is the number of denoising steps.}
  \label{tab:abl_fkl}
  \begin{tabular}{@{}l cccc@{}}
    \toprule
    Configuration & Quality$\uparrow$ & Semantic$\uparrow$ & Dynamic$\uparrow$ & Drift$\downarrow$ \\
    \midrule
    bi-attn + 1 step (default)  & \textbf{0.846} & \textbf{0.723} & 0.686 & \textbf{0.257} \\
    causal + 1 step             & 0.828 & 0.701 & 0.625 & 0.271 \\
    bi-attn + 2 steps           & 0.835 & 0.708 & \textbf{0.693} & 0.296 \\
    bi-attn + 4 steps           & 0.813 & 0.684 & 0.691 & 0.306 \\
    \midrule
    w/o $\mathcal{L}_{\text{FKL}}$        & 0.839 & 0.732 & 0.445 & 0.218 \\
    w/o re-training             & 0.767 & 0.559 & 0.512 & 0.309 \\
    w/o hier.\ denoising (Self-Forcing)  & 0.829 & 0.708 & 0.542 & 0.355 \\
    \bottomrule
  \end{tabular}
  \vspace{-2mm}
\end{table}

\begin{figure}[tb]
    \centering
    \includegraphics[width=.65\linewidth]{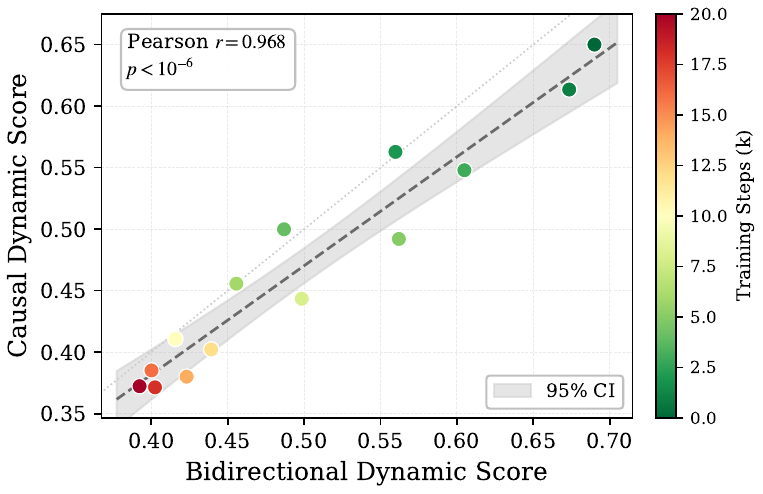}
    \caption{\textbf{Correlation between bidirectional and causal dynamics during training (w/o $\mathcal{L}_{\text{FKL}}$).}
    Each point represents one training checkpoint; colour encodes the training step.
    A strong positive correlation (Pearson $r=0.968$) confirms that the low-motion shortcut affects both attention modes simultaneously and that regularising the bidirectional mode effectively constrains causal-mode dynamics.
    }
    \label{fig:dynamics_corr}
    \vspace{-1mm}
\end{figure}

\begin{figure}[t]
    \centering
    \includegraphics[width=1.0\linewidth]{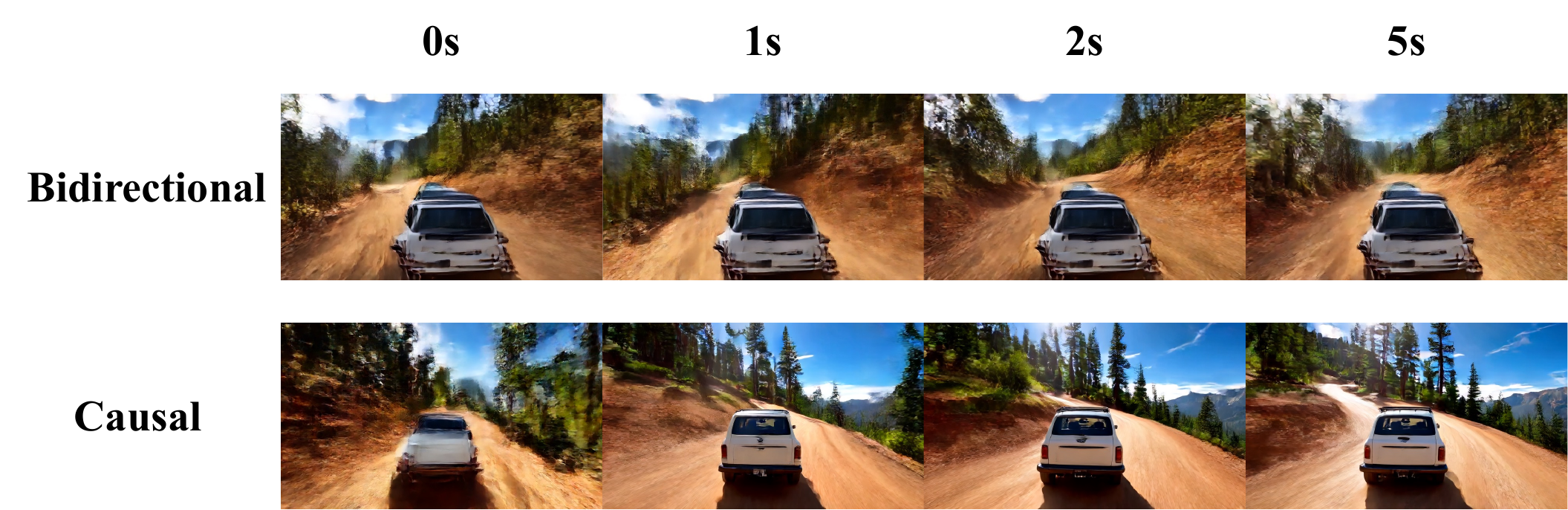}
    \vspace{-3mm}
    \caption{\textbf{Comparison of single-step denoising under bidirectional vs.\ causal attention.}
    Bidirectional attention produces frames of uniform quality and blur across all positions, while causal attention yields progressively sharper frames as preceding context reduces uncertainty for later positions.
    }
    \label{fig:bi-caus}
\end{figure}

\emph{Attention mode.}
Applying $\mathcal{L}_{\text{FKL}}$ in causal mode (``causal + 1 step'') leads to lower dynamics (0.625 vs.\ 0.686) and reduced quality compared with the bidirectional-attention default.
To empirically justify our design of applying $\mathcal{L}_{\text{FKL}}$ in bidirectional-attention mode, we track the dynamic scores under both attention modes across training checkpoints (without $\mathcal{L}_{\text{FKL}}$).
As shown in Fig.~\ref{fig:dynamics_corr}, both modes exhibit a consistent decline in dynamics over training, and the two scores are strongly positively correlated (Pearson $r=0.968$).
This confirms that regularising the bidirectional mode serves as an effective and non-intrusive proxy for preserving causal-mode motion diversity.
Fig.~\ref{fig:bi-caus} visualises single-step denoising outputs under both modes.
Under bidirectional attention, all frames exhibit a uniform level of quality and blur, since the full-sequence attention treats every position symmetrically.
In contrast, causal denoising produces frames that become progressively sharper along the temporal axis: as preceding frames fix the low-frequency structure, the conditional distribution of later frames concentrates, resulting in higher-frequency details.
This asymmetry means that a distillation target derived from bidirectional denoising provides a spatiotemporally uniform supervision signal well suited to regularising global dynamics, whereas directly constraining causal outputs introduces mismatched targets that are tightly coupled with the model's autoregressive generation pathway, degrading overall quality.
Bidirectional-mode regularisation is therefore the preferred configuration.

\emph{Number of constrained steps.}
Increasing $K$ from 1 to 2 or 4 brings marginal gains in dynamics (0.693, 0.691 vs.\ 0.686) but monotonically degrades both quality and drift.
This confirms that motion diversity is primarily governed by the low-frequency structure laid down in the first denoising step; constraining subsequent high-frequency refinement steps provides diminishing returns while interfering with the model's denoising capacity.
A single constrained step ($K=1$) is therefore sufficient and optimal.

\emph{Component necessity.}
Removing $\mathcal{L}_{\text{FKL}}$ entirely (``w/o $\mathcal{L}_{\text{FKL}}$'') yields competitive quality and the lowest drift, but dynamics collapse drastically (0.445), confirming that the model falls into the low-motion shortcut without forward-KL regulation.
``w/o re-training'' applies hierarchical denoising only at inference without corresponding training, which significantly reduces drift compared with Self-Forcing (0.309 vs.\ 0.355) yet at a substantial cost to visual quality (Quality 0.767), highlighting the importance of train--test alignment.
Finally, removing hierarchical denoising altogether recovers the standard Self-Forcing baseline, which exhibits the highest drift (0.355) and lower dynamics (0.542), validating the contribution of hierarchical denoising to long-horizon stability.

\section{Conclusion}

We presented \textbf{HiAR}, a hierarchical denoising framework that addresses the distribution drift problem in autoregressive long video generation.
Our key insight is that a fully clean context is unnecessary and, in fact, harmful: by conditioning each block on context at matched noise level rather than predicted clean frames, hierarchical denoising attenuates inter-block error propagation while preserving temporal causality.
This simple reordering---from the conventional block-first pipeline to a step-first paradigm---also enables pipelined parallel inference, achieving $\sim$1.8$\times$ wall-clock speedup in our 4-step setting.
To stabilise training, we introduced a forward-KL regulariser in bidirectional-attention mode that counteracts the low-motion shortcut inherent to reverse-KL distillation, preserving motion diversity without interfering with the DMD objective.
Experiments on VBench and a dedicated drift metric confirm that HiAR achieves the best overall quality and the lowest temporal degradation among all compared methods on 20-second generation.

\bibliographystyle{plainnat}
\bibliography{main}
\end{document}